\documentclass[runningheads]{llncs}
\usepackage{graphicx}
\usepackage{amsfonts,amsmath,amssymb}
\usepackage{mathrsfs}
\usepackage{graphicx}
\usepackage{subfig}
\usepackage{booktabs}
\usepackage{dblfloatfix} 
\usepackage{array}
\newcolumntype{I}{!{\vrule width 3pt}}
\newlength\savedwidth

\begin{document}
%
\title{Unsupervised Wasserstein Distance Guided Domain Adaptation for 3D Multi-Domain Liver Segmentation}

%
\titlerunning{Unsupervised Wasserstein Distance Guided Domain Adaption}
%


\author{Chenyu You\inst{1}$^\star$ \and
Junlin Yang\inst{2}$^\star$ \and
Julius Chapiro\inst{3} \and
James S. Duncan\inst{1,2,3,4}}

%
\authorrunning{C. You et al.}
%
\institute{Department of Electrical Engineering, Yale University, New Haven, CT, USA \\\email{chenyu.you@yale.edu} \and
Department of Biomedical Engineering, Yale University, New Haven, CT, USA \and
Department of Radiology \& Biomedical Imaging, Yale School of Medicine, New Haven, CT, USA \and
Department of Statistics \& Data Science, Yale University, New Haven, CT, USA}
%
\maketitle              
%

\begin{abstract}
Deep neural networks have shown exceptional learning capability and generalizability in the source domain when massive labeled data is provided. However, the well-trained models often fail in the target domain due to the domain shift. Unsupervised domain adaptation aims to improve network performance when applying robust models trained on medical images from source domains to a new target domain. In this work, we present an approach based on the Wasserstein distance guided disentangled representation to achieve 3D multi-domain liver segmentation. Concretely, we embed images onto a shared content space capturing shared feature-level information across domains and domain-specific appearance spaces. The existing mutual information-based representation learning approaches often fail to capture complete representations in multi-domain medical imaging tasks. To mitigate these issues, we utilize Wasserstein distance to learn more complete representation, and introduces a content discriminator to further facilitate the representation disentanglement. Experiments demonstrate that our method outperforms the state-of-the-art on the multi-modality liver segmentation task.
\let\thefootnote\relax\footnote{$^\star$~Equal contribution}
\let\thefootnote\relax\footnote{This work was supported by NIH Grant 5R01 CA206180}
\end{abstract}
%
%
%

\section{Introduction}


Accurate and consistent measurements on medical images greatly assist radiologists in making precise and reliable diagnoses and staging the patients. In clinical practices, manual segmentation of anatomical structures from 3D medical images by experienced experts is tedious, time-consuming, and error-prone, which is not suitable for large-scale studies~\cite{greenspan2016guest}. Besides, different medical imaging modalities, such as Magnetic Resonance Imaging (MRI), Computed Tomography (CT), and Positron Emission Tomography (PET), provide unique views of tissue features at different spatial resolutions with functional information. In particular, CT is the most common imaging modality for the diagnosis of hepatocellular carcinoma (HCC), the primary malignant tumor in the human liver. However, the scan is associated with the radiation dosage and provides low soft-tissue contrast, which makes it difficult to visualize tumor boundaries. As a non-invasive technique, MRI offers higher contrast, but has disadvantages in assessment cost, acquisition time, and is more prone to artifacts. In clinical practice, the fusion of multi-modal images allows for capturing more anatomical information and integrating complementary information to minimize redundancy and enhancing the diagnostic potential. Thus, it is a rapidly rising demand to segment cross-modality images for accurate analysis and interpretation.



\begin{figure}[!t]
\begin{center}
\includegraphics[width=0.9\linewidth]{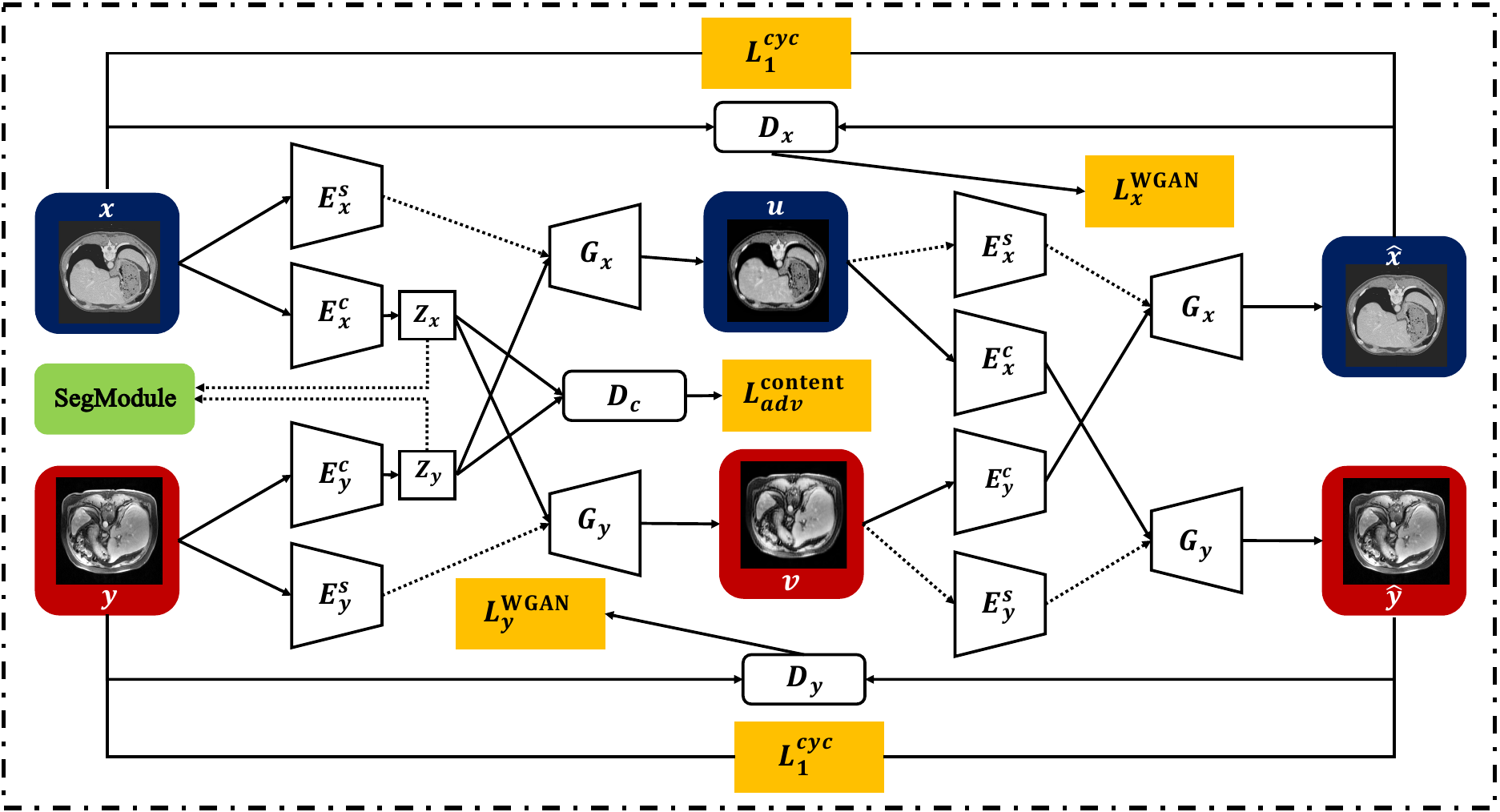}
\end{center}
\caption{Overview of the proposed $3$D Wasserstein Distance Guided Domain Adaption Model. With the guidance of the following constraints~$\mathcal{L}^{\mathrm{WGAN}}$,~$\mathcal{L}^{\mathrm{cyc}}$ and~$\mathcal{L}^{\mathrm{content}}$, we can learn the cross-domain mapping between unpaired CT and Multi-phasic MRI sequences. The domain discriminators~$\{D_{\mathcal{X}},D_{\mathcal{Y}}\}$, and a content discriminator~$D^c$ jointly encourage the model to obtain the well-learned representation disentanglement.}
\label{fig2}
\end{figure}


Unsupervised domain adaptation has been widely used for generalizing medical image segmentation models across domains. The major challenge is to mitigate domain gaps between different modalities. Several recent efforts have been made to improve the segmentation performance without label data in the medical imaging community~\cite{dong2018unsupervised,ouyang2019data,yang2019unsupervised}. For example, Yang~\textit{et al.}~\cite{yang2019unsupervised} utilized multi-modal unsupervised image-to-image translation framework (MUNIT)~\cite{huang2018multimodal} to decompose image into a shared domain-invariant content space and a domain-specific style space. Then, the learned content representations are used to train the segmentation network.



In this paper, we present a novel unsupervised cross-modality domain adaptation method for medical image segmentation. Our proposed method extends upon~\cite{yang2019unsupervised} as follows: Firstly, in order to obtain more complete domain invariant representations, we introduce Wasserstein distance~\cite{gulrajani2017improved,you2018structurally,you2019low} to reduce the domain discrepancy instead of the negative log-likelihood used in~\cite{lee2018diverse}. Secondly, medical imaging data is inherently three-dimensional (3D). However, most of domain adaption methods leverage 2D information. We incorporate 3D volumetric information to improve the image quality of reconstructed images by fully exploiting detailed spatial information along the $z$ dimension. Thirdly, to facilitate the decomposition of domain-invariant shared information and domain-specific features, in our work we propose a content discriminator to distinguish extracted content-level representations between different domains and utilize a cross-cycle consistency loss to enforce many-to-many mappings. We demonstrate that our proposed methods are competitive against other state-of-the-art methods in multi-domain liver segmentation.




\section{Method}
\label{sec:method}
\subsection{Overview}
Our goal is to learn a cross-modality mapping between two domains $\mathcal{X}$ and $\mathcal{Y}$ without paired training data. We assume that there exists a potentially many-to-many mapping between two domains. Our approach decomposes images onto a shared content spaces $c\in\mathcal{C}$, and domain-specific space $\mathcal{S}_\mathcal{X}$ and $\mathcal{S}_\mathcal{Y}$~\cite{huang2018multimodal,lee2018diverse}. Intuitively, the content encoders are used to map the shared information shared among domains onto $\mathcal{C}$, and the style encoders project the domain-specific information onto $\mathcal{S}_\mathcal{X}$ and $\mathcal{S}_\mathcal{Y}$.


\subsection{Model}
Let $x\in\mathcal{X}$ and $y\in\mathcal{Y}$ be images from two different domains, or in our task, two different imaging modalities. As shown in Fig.~\ref{fig2}, Our method deploys 3D CNN to take advantage of spatial information. Similar to the recent works~\cite{huang2018multimodal,lee2018diverse}, the overall model consists of
several networks: jointly trained content encoders~\{$E_{\mathcal{X}}^{c}$,$E_{\mathcal{Y}}^{c}$\}, style encoders $\{E_{\mathcal{X}}^{s}$,$E_{\mathcal{Y}}^{s}\}$, decoders $\{G_{\mathcal{X}}$,$G_{\mathcal{Y}}\}$ and domain discriminators $\{D_{\mathcal{X}},D_{\mathcal{Y}}\}$, and a content discriminator~$D^c$.~\textit{i.e.}, given the domain~$\mathcal{X}$, the content encoder $E_{\mathcal{X}}^{c}$ and the style encoder $E_\mathcal{X}^{s}$ encode~$x$ to a content code~$z_{x}^{c}$ in a shared, domain-invariant content space ($\mathcal{C}$) and a style code~$z_{x}^{s}$ in domain-specific style space ($\mathcal{S}_{\mathcal{X}}$), respectively. The decoders~$G_{\mathcal{X}}$ reconstruct images conditioned on both content and style codes. The discriminator~$D_\mathcal{X}$ aims to discriminate between real images and reconstructed images from the domain $\mathcal{X}$. In addition, the content discriminator~$D^{c}$ is trained jointly to distinguish the encoded content features~$z_{x}^{c}$ and~$z_{y}^{c}$ between two domain.

\paragraph{\textbf{Latent reconstruction.}}To achieve representation disentanglement and preserve maximal information in the representation of each domain, we use the bidirectional reconstruction loss to encourage the bidirectional mapping which includes the self-reconstruction loss and latent reconstruction loss,~\textit{i.e.},
\begin{subequations}
    \begin{align}
        \mathcal{L}_{\mathcal{X}}^{\mathrm{recon}} &= {\mathbb{E}_{x}}[||G_{\mathcal{X}}(E_{\mathcal{X}}^{c}(x),E_{\mathcal{X}}^{s}(x))-x||_{1}], \\
        \mathcal{L}_{\mathcal{X}^c}^{\mathrm{latent}} &= {\mathbb{E}_{x,y}}[||E_{\mathcal{Y}}^c(G_{\mathcal{Y}}(z_{x}^{c},z_{y}^{s}))-z_{x}^c||_{1}], \\
        \mathcal{L}_{\mathcal{Y}^s}^{\mathrm{latent}} &= {\mathbb{E}_{x,y}}[||E_{\mathcal{Y}}^s(G_{\mathcal{Y}}(z_{x}^{c},z_{y}^{s}))-z_{y}^s||_{1}],
    \end{align}
\end{subequations}
\paragraph{\textbf{In domain reconstruction.}}In order to facilitate the disentangled content and attribute representations for cyclic reconstruction, we formulate the cross-cycle consistency loss as~\cite{lee2018diverse}:
\begin{multline}
    \mathcal{L}^{\mathrm{cyc}} = \mathcal{L}_{\mathcal{X}}^{\mathrm{cyc}} + \mathcal{L}_{\mathcal{Y}}^{\mathrm{cyc}} = {\mathbb{E}_{x,y}}[||G_{\mathcal{X}}(E_{\mathcal{Y}}^{c}(\hat x),E_{\mathcal{X}}^{s}(\hat y))-x||_{1} \\
    + ||G_{\mathcal{Y}}(E_{\mathcal{X}}^{c}(\hat y),E_{\mathcal{Y}}^{s}(\hat x))-y||_{1}],
\end{multline}
where $\hat y = G_{\mathcal{X}}(E_{\mathcal{Y}}^c(y),E_{\mathcal{X}}^s(x))$ and $\hat x = G_{\mathcal{Y}}(E_{\mathcal{X}}^c(x),E_{\mathcal{Y}}^s(y))$, respectively.





\paragraph{\textbf{Adversarial Loss.}}First, we introduce WGAN-GP~\cite{gulrajani2017improved,you2019ct} to match the distribution of reconstructed images to the target domain. The generator $G$ and two discriminators $D_{\mathcal{X}}$ and $D_{\mathcal{Y}}$ are trained via alternatively optimizing the corresponding composite loss functions.~\textit{i.e.}:
\begin{equation}
\mathcal{L}_{\mathcal{Y}}^{\mathrm{WGAN}}={\mathbb{E}_{y}}[D_{\mathcal{Y}}(y)] - {\mathbb{E}_{\hat{y}}}[D_{\mathcal{Y}}(\hat{y})] +\alpha\cdot L_{\mathcal{Y}}^{grad},
\end{equation}
where $D_{\mathcal{Y}}$ is a discriminator for domain adaption to distinguish between reconstructed images~$\hat{y}$ and real images~$y$. The coefficient $\alpha$ is a weighting hyperparameter. The gradient penalty term is~$L_{\mathcal{Y}}^{grad} = \mathbb{E}_{\tilde{y}}[(||\nabla_{\tilde{y}}D_{\mathcal{Y}}(\tilde{y})-1||_2)^2)]$, where $\tilde{y}$ is uniformly sampled between $y$ and $\hat{y}$. The discriminator $D_{\mathcal{X}}$ and loss $\mathcal{L}_{\mathcal{X}}^{\mathrm{WGAN}}$ are defined similarly. Second, we employ a content discriminator $D_{\mathcal{C}}$ to match the distribution of the encoded content features $z_x$ and $z_y$ of different domains. We formulate the content adversarial loss~\cite{lee2018diverse} as:
\begin{multline}
\mathcal{L}^{\mathrm{content}}(E_{\mathcal{X}}^{c},E_{\mathcal{Y}}^{c},D_{\mathcal{C}}) = \mathop{\min}_{G}\mathop{\max}_{D}~\mathbb{E}_{x}[\frac{1}{2}\log D_{\mathcal{C}}(E_{\mathcal{X}}^{c}(x))+\frac{1}{2}\log(1-D_{\mathcal{C}}(E_{\mathcal{X}}^{c}(x)))] \\
+ \mathbb{E}_{y}[\frac{1}{2}\log D_{\mathcal{C}}(E_{\mathcal{Y}}^{c}(y))+\frac{1}{2}\log(1-D_{\mathcal{C}}(E_{\mathcal{Y}}^{c}(y)))]
\end{multline}
\paragraph{\textbf{Total Loss.}} We jointly train the encoders, decoders, and discriminators via optimizing the following objective function.
\begin{multline}
\mathop{\min}_{G_{\mathcal{X}},G_{\mathcal{Y}},E_{\mathcal{X}},E_{\mathcal{Y}}}\mathop{\max}_{D_{\mathcal{X}},D_{\mathcal{Y}},D_{\mathcal{C}}}\mathcal{L}(G_{\mathcal{X}},G_{\mathcal{Y}},E_{\mathcal{X}},E_{\mathcal{Y}},D_{\mathcal{X}},D_{\mathcal{Y}},D_{\mathcal{C}}) = \lambda^{\mathrm{WGAN}} \mathcal{L}^{\mathrm{WGAN}} \\ 
+ \lambda^{\mathrm{recon}} \mathcal{L}^{\mathrm{recon}}
+ \lambda^{\mathrm{cyc}} \mathcal{L}^{\mathrm{cyc}} +\lambda^{\mathrm{latent}} \mathcal{L}^{\mathrm{latent}} + \lambda^{\mathrm{content}}\mathcal{L}^{\mathrm{content}},
\end{multline}
where $\lambda^{\mathrm{WGAN}},\lambda^{\mathrm{recon}},\lambda^{\mathrm{cyc}},\lambda^{\mathrm{latent}},\lambda^{\mathrm{content}}$ are weights that control the importance of each term.


\subsubsection{SegModule}
Once disentangled representation is achieved, the content-only image can be generated given the content code. For both CT and MRI, we assume that their content codes are embedded onto the shared domain-invariant latent space that preserve anatomical information but exclude modality-specific information. We implement DenseNet~\cite{huang2017densely} as the segmentation network. Note that we tailored the network configuration for our task. To address the inherent class imbalance between foreground liver part and the background, we combine the Soft Dice and weighted Cross-Entropy (CE) losses~\cite{ouyang2019data} to train the SegModule.

\subsubsection{Model Implementation}
We implement the proposed method in PyTorch, using NVIDA TITAN XP GPUs. For domain adaption tasks, we build our model based on~\cite{huang2018multimodal,lee2018diverse} with changes as discussed in section~\ref{sec:method}. The network architecture here includes a VAE with two domain-specific encoders and decoders that is based on~\cite{huang2018multimodal}. We utilize the Wasserstein distance with gradient penalty instead of the negative log-likelihood. The content discriminator adopts the same architecture as in~\cite{lee2018diverse}. We use Adam optimizer with a learning rate of $10^{-4}$ and set the hyperparameter~$\lambda^{\mathrm{WGAN}},\lambda^{\mathrm{recon}},\lambda^{\mathrm{cyc}},\lambda^{\mathrm{latent}},\lambda^{\mathrm{content}}$, and $\alpha$ as $1.0,10.0,0.1,10$, and $10.0$.
The content discriminator is updated every $3$ iterations. At the rest iterations, other discriminators and generators would be updated jointly, leveraging the advantage of content discriminator to align the content code across different domains.



\begin{table*}[!ht]
\renewcommand{\arraystretch}{0.9}
\centering
\caption{Comparison over Domain Adaptation.}
\label{table:eg1}
\setlength{\tabcolsep}{10pt}
\begin{tabular}{c c c}
\hline\hline
Method & Dice & Jaccard \\
\cline{1-3}
DenseNet$^s$~\cite{huang2017densely} & 0.362$\pm$0.016 & 0.325$\pm$0.047 \\
CycleGAN~\cite{CycleGAN2017} & 0.753$\pm$0.031 & 0.681$\pm$0.083 \\
DADR~\cite{yang2019unsupervised} & 0.828$\pm$0.072 & 0.757$\pm$0.092 \\
3D-WDGDA & 0.837$\pm$0.054 & 0.759$\pm$0.065 \\
3D-WDGDA$^{c}$ & \textbf{0.875$\pm$0.039} & \textbf{0.814$\pm$0.027}\\
\hline\hline
\end{tabular}
\label{table:eg2}
\end{table*}

\begin{figure*}[!t]
\centering
\captionsetup[subfigure]{labelformat=empty}
\subfloat[Ground-Truth]{\includegraphics[width=0.8in]{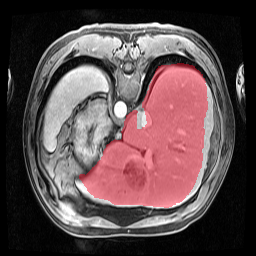}\label{fig: SR1}}
\subfloat[DenseNet$^{s}$]{\includegraphics[width=0.8in]{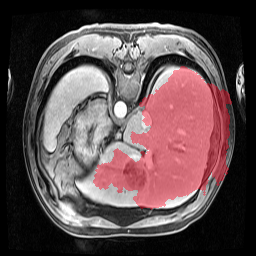}\label{fig: FSRCNN1}}
\subfloat[CycleGAN]{\includegraphics[width=0.8in]{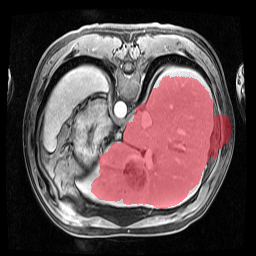}\label{fig: FSRCNN1}}
\subfloat[DADR]{\includegraphics[width=0.8in]{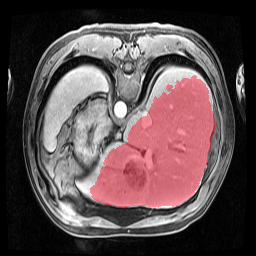}\label{fig: LR1}}
\subfloat[3D-WDGDA]{\includegraphics[width=0.8in]{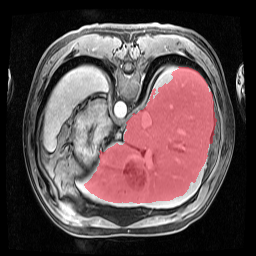}\label{fig: aplus}}
\subfloat[3D-WDGDA$^{c}$]{\includegraphics[width=0.8in]{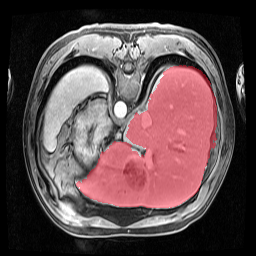}\label{fig: FSRCNN1}}



\subfloat[Ground-Truth]{\includegraphics[width=0.8in]{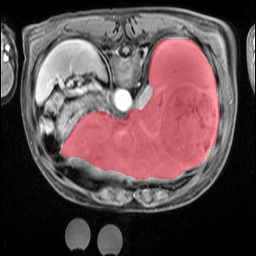}\label{fig: SR1}}
\subfloat[DenseNet$^{s}$]{\includegraphics[width=0.8in]{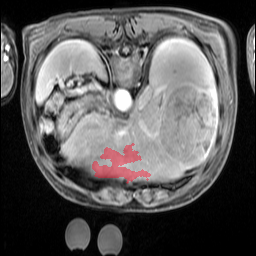}\label{fig: FSRCNN1}}
\subfloat[CycleGAN]{\includegraphics[width=0.8in]{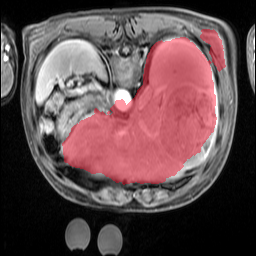}\label{fig: FSRCNN1}}
\subfloat[DADR]{\includegraphics[width=0.8in]{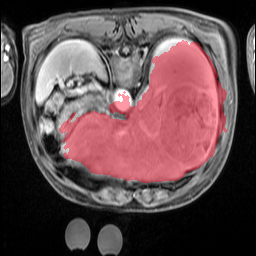}\label{fig: LR1}}
\subfloat[3D-WDGDA]{\includegraphics[width=0.8in]{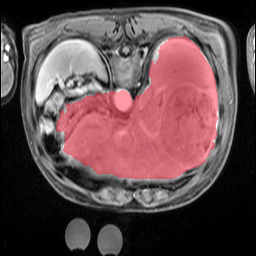}\label{fig: aplus}}
\subfloat[3D-WDGDA$^{c}$]{\includegraphics[width=0.8in]{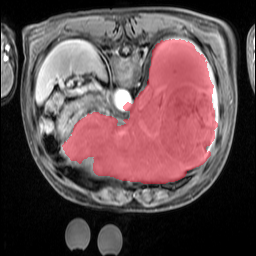}\label{fig: FSRCNN1}}

\caption{Qualitative results of different methods on segmentation. We list the ground-truth, DenseNet$^{s}$, CycleGAN, DADR, 3D-WDGDA, 3D-WDGDA$^{c}$ (with content discriminater).}
\label{fig:eg1}
\end{figure*}

\section{Experiments}
\subsection{Datasets and training settings}
We used two datasets for validation: 1).~\textrm{LiTS - Liver Tumor Segmentation Challenge} dataset~\cite{christ2017lits}. It consists of $131$ contrast-enhanced 3D abdominal CT scans. 2). Multi-phasic MRI scans of $36$ local patients with HCC (note that the CT and MRI scans are~\textbf{unpaired} and~\textbf{unmatched}). Considering the clinical practise, we chose CT scans as source domain and MRI scans as target domain. We use 5-fold cross validation on the CT and MRI datasets, and normalized as zero mean and unit variance. In both WDGDA and SegModule part, input size of $3$D modules is $256\times256\times5$, and for $2$D modules is $256\times256$. To avoid over-fitting, we used standard data augmentation methods, including randomly flipping and rotating along the axial plane.
We evaluated two variations of the proposed method: our proposed 3D Wasserstein Distance Guided Domain Adaptation model without content discriminator (3D-WDGDA), and 3D-WDGDA with content discriminator (3D-WDGDA$^{c}$). 


In this work, there are three experiment setups: 1). For the domain adaption part, we use $4$ folds of CT and $4$ folds of pre-contrast MRI for training. Then $4$ folds of content-only CT and $1$ fold of pre-contrast MRI are used to train and test the SegModule, respectively. 2). We follow the same domain adaption setting in experiment $1$, then utilize $4$ folds of content-only CT and $4$ fold of pre-contrast MRI as network input to train the SegModule, and $1$ fold of pre-contrast MRI as test dataset. 3). $4$ folds of CT and $4$ folds of multi-phasic MRI are used for training. For the segmentation part, we investigate the multi-modal target domain by using $4$ folds of CT and $4$ folds of multi-phasic MRI. Note that multi-phasic MRIs themselves are multi-modal target domain since they contain several MRI modalities. We evaluate segmentation performance in terms of two metrics: Dice and Jaccard.

\subsection{Results}
\subsubsection{Experiment $1$:} To demonstrate the domain shift problem, we first evaluate the performance of the unadpated baseline by directly feeding target images to DenseNet$^{s}$~\cite{huang2017densely}. We further compare our methods with CycleGAN$+$DenseNet, DADR~\cite{yang2019unsupervised}. We present two typical results in Fig.~\ref{fig:eg1}. The quantitative results are shown in Table~\ref{table:eg1}. Compared with other methods, the proposed 3D-WDGDA$^{c}$ improves the segmentation performance, and achieves an average Dice of $0.875$ and Jaccard of $0.814$.\vspace{-10pt}

\begin{table}[!t]
\renewcommand{\arraystretch}{0.9}
\centering
\caption{Comparison over Joint-Domain.}
\setlength{\tabcolsep}{5pt}
\begin{tabular}{c c c c c c}
\hline\hline
& \multicolumn{2}{c}{CT} && \multicolumn{2}{c}{MRI}\\
& Dice & Jaccard && Dice & Jaccard\\
\cline{2-3}\cline{5-6}
DenseNet$^s$~\cite{huang2017densely} & 0.807$\pm$0.031 & 0.793$\pm$0.035 && 0.821$\pm$0.017 & 0.722$\pm$0.028\\
DADR~\cite{yang2019unsupervised} & 0.811$\pm$0.076 & 0.780$\pm$0.067 && 0.828$\pm$0.022 & 0.727$\pm$0.049\\
3D-WDGDA & 0.885$\pm$0.026 & 0.801$\pm$0.058 && 0.843$\pm$0.047 & 0.735$\pm$0.045\\
3D-WDGDA$^{c}$ & \textbf{0.904$\pm$0.012} & \textbf{0.831$\pm$0.041} && \textbf{0.883$\pm$0.036} & \textbf{0.802$\pm$0.038}\\
\hline\hline
\end{tabular}
\end{table}


\begin{figure*}[!t]
\centering
\captionsetup[subfigure]{labelformat=empty}
\subfloat[CT]{\includegraphics[width=0.90in]{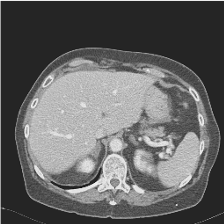}\label{fig: SR1}}
\subfloat[DADR]{\includegraphics[width=0.90in]{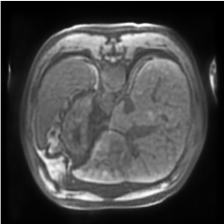}\label{fig: LR1}}
\subfloat[3D-WDGDA]{\includegraphics[width=0.90in]{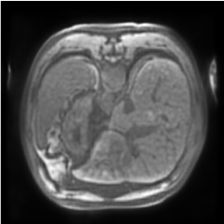}\label{fig: aplus}}
\subfloat[3D-WDGDA$^{c}$]{\includegraphics[width=0.90in]{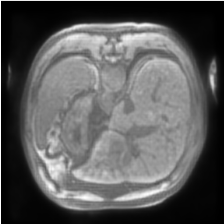}\label{fig: FSRCNN1}}
\subfloat[MRI]{\includegraphics[width=0.90in]{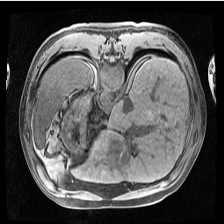}\label{fig: FSRCNN1}}



\subfloat[CT]{\includegraphics[width=0.90in]{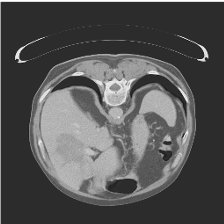}\label{fig: SR1}}
\subfloat[DADR]{\includegraphics[width=0.90in]{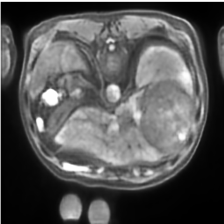}\label{fig: LR1}}
\subfloat[3D-WDGDA]{\includegraphics[width=0.90in]{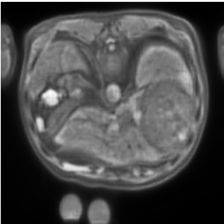}\label{fig: aplus}}
\subfloat[3D-WDGDA$^{c}$]{\includegraphics[width=0.90in]{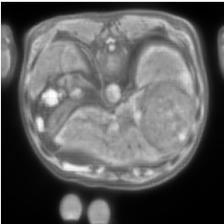}\label{fig: FSRCNN1}}
\subfloat[MRI]{\includegraphics[width=0.90in]{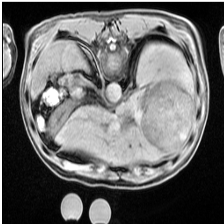}\label{fig: FSRCNN1}}

\caption{Visualization of content-only images by different methods. We list the CT images, DADR, 3D-WDGDA, 3D-WDGDA$^{c}$, and the reference MRI images.}
\label{fig:eg2}
\end{figure*}



\subsubsection{Experiment $2$:} To show the robustness of our method for joint training, we compare our methods with other state-of-the-art methods. As shown in Table.~\ref{table:eg2}, our proposed method 3D-WDGDA$^{c}$ consistently obtains the highest Dice and Jaccard score over CT and MRI datasets. Visual results of the proposed 3D-WDGDA$^{c}$ are shown in Fig.~\ref{fig:eg2}.

\begin{table*}[!ht]
\renewcommand{\arraystretch}{0.9}
\centering
\caption{Comparison over Multi-modal Target Domain. For brevity, CT$\rightarrow$MRI denotes that SegModule is trained with with content-only CT images and tested by multi-phasic MRI images.}
\setlength{\tabcolsep}{5pt}
\begin{tabular}{c c c c c c}
\hline\hline
& \multicolumn{2}{c}{CT$\rightarrow$MRI} && \multicolumn{2}{c}{CT$\rightarrow$CT}\\
& Dice & Jaccard && Dice & Jaccard \\
\cline{2-3}\cline{5-6}
DenseNet$^s$~\cite{huang2017densely} & 0.469$\pm$0.005 & 0.289$\pm$0.004 && 0.896$\pm$0.048 & 0.821$\pm$0.002 \\
DADR~\cite{yang2019unsupervised} & 0.736$\pm$0.034 & 0.619$\pm$0.059 && 0.893$\pm$0.038 & 0.824$\pm$0.048 \\
3D-WDGDA & 0.776$\pm$0.013 & 0.677$\pm$0.053 && 0.902$\pm$0.056 & 0.832$\pm$0.037 \\
3D-WDGDA$^{c}$ & \textbf{0.834$\pm$0.029} & \textbf{0.707$\pm$0.047} && \textbf{0.919$\pm$0.044} & \textbf{0.851$\pm$0.053} \\
\hline\hline
& \multicolumn{2}{c}{MRI$\rightarrow$CT} && \multicolumn{2}{c}{MRI$\rightarrow$MRI}\\
& Dice & Jaccard && Dice & Jaccard \\
\cline{2-3}\cline{5-6}
DenseNet$^s$~\cite{huang2017densely} & 0.766$\pm$0.003 & 0.631$\pm$0.038 && 0.851$\pm$0.015 & 0.725$\pm$0.016 \\
DADR~\cite{yang2019unsupervised} & 0.782$\pm$0.019 & 0.674$\pm$0.015 && 0.854$\pm$0.022 & 0.739$\pm$0.031 \\
3D-WDGDA & 0.796$\pm$0.016 & 0.718$\pm$0.035 && 0.869$\pm$0.047 & 0.740$\pm$0.064 \\
3D-WDGDA$^{c}$ & \textbf{0.807$\pm$0.044} & \textbf{0.744$\pm$0.057} && \textbf{0.881$\pm$0.027} & \textbf{0.786$\pm$0.031} \\
\hline\hline
\end{tabular}
\label{table:eg3}
\end{table*}

\subsubsection{Experiment $3$:} Multi-phasic MRI are considered as multi-modal target domain with complex statistics. We therefore analyze the effectiveness of the proposed method in multi-modal target domain. The quantitative results are shown in Table.~\ref{table:eg3}.~\textit{i.e.}, for brevity, CT$\rightarrow$MRI denotes that SegModule is trained with content-only CT images and tested by multi-phasic MRI images. We can see that our method clearly remains effective with the multi-modal target domain.

\section{Conclusions and Discussions}
We present a novel $3$D unsupervised cross-modality Wasserstein distance guided domain adaptation method for medical image segmentation, which would improve clinical decision support systems by leveraging unpaired multi-parametric MRI and CT data. Our method applies Wasserstein distance for the adversarial training, and further takes advantage of~$3$D~CNN to capture spatial information. More importantly, we introduce a content discriminator to encourage content features not to carry modality-specific information, and further preserve feature-level anatomical information for the segmentation task. Qualitative and quantitative results demonstrate the superiority of proposed model over the multi-modal image reconstruction in clinical domains, which is consistent with quantitative evaluations in terms of traditional image segmentation measures. Future work includes improving the efficiency of the proposed methods.
\bibliographystyle{splncs04}
\bibliography{reference}
%




\end{document}